\documentclass{article}
\usepackage{iclr2027_conference,times}

\usepackage[utf8]{inputenc}
\usepackage[T1]{fontenc}
\usepackage{graphicx}
\usepackage{booktabs}
\usepackage{multirow}
\usepackage{xcolor}
\usepackage[font=small,labelfont=bf]{caption}
\usepackage{tikz}
\usetikzlibrary{arrows.meta,positioning,shapes.geometric,calc,backgrounds,fit}
\usepackage{amsmath}
\usepackage{amssymb}
\usepackage{listings}
\usepackage{url}
\usepackage{hyperref}
\usepackage{enumitem}
\setlist{nosep,leftmargin=1.4em}
\definecolor{navy}{HTML}{003C71}
\definecolor{acc}{HTML}{E8821E}
\definecolor{grn}{HTML}{1E8E5A}
\definecolor{codebg}{HTML}{F4F6F8}

\title{Improving VLMs on scientific chart analysis via autonomous evolution}

\author{%
\rule{0pt}{75pt}%
\hspace*{-\tabcolsep}%
\makebox[\textwidth][c]{\normalfont
\begin{tabular}{@{}c@{\hspace{3em}}c@{}}
Abhinav Sudhakar Dubey & Vidushi Goel \\
University of California, Santa Cruz & University of California, Santa Cruz \\
\texttt{adubey4@ucsc.edu} & \texttt{vgoel5@ucsc.edu} \\[1.1em]
Kiran Rana & Shikhar Mathur \\
University of California, Santa Cruz & Airbnb Inc. \\
\texttt{kirana@ucsc.edu} & \texttt{shikharmathur96@gmail.com} \\[1.1em]
\multicolumn{2}{c}{Vaggos Chatziafratis} \\
\multicolumn{2}{c}{University of California, Santa Cruz} \\
\multicolumn{2}{c}{\texttt{vaggos@ucsc.edu}}
\end{tabular}}%
\hspace*{-\tabcolsep}%
}

\iclrfinalcopy
\begin{document}

\maketitle
\lhead{}
\renewcommand{\headrulewidth}{0pt}

\begin{abstract}
Understanding and operating with scientific charts is a standard approach in data analysis, and while humans are good at interpreting such charts, open vision-language models (VLMs) are known to perform poorly when it comes to question answering based on such data types. Motivated by an initial observation that many of their errors are due to inexact answering, we raise the following natural question: \textit{how much accuracy can we recover without changing the model at all?} In this paper, we study this question by searching over inference programs and the code around a ``frozen'' version of the model. Using ideas from autonomous evolution (famously used in AlphaEvolve), we first start from a baseline inference program, and in each round we show an LLM judge the code together with a few failures on training questions, and ask for an improving evolutionary move. As usual in autonomous evolution approaches, the candidate programs are kept in a small pool ranked by training accuracy, and the next edit is performed on top-scorers from this pool. We run this approach on three VLMs and four popular benchmarks on scientific charts, and we measure the fraction of descriptive questions whose answer exactly matches the ground truth. The accuracy improves in all experimental settings. On average, it rises from 16.6\% with the starting program to 41.3\% with the evolved one, and the largest single increase is 55.2 percentage points. As a side benefit, our evolved program is 3.8$\times$ faster than the starting program, because short answers are easier to grade and faster to generate. Our results demonstrate that a large part of the observed errors arise due to the inference pipeline, and that optimizing it can improve the ``frozen'' VLM without the need to retrain it.
\end{abstract}

\section{Introduction}
\label{sec:intro}

Plots are one of the key mediums by which scientific results are often reported, and analyzing them is an increasingly common task that is being given as input to vision-language models (VLMs). Unfortunately, while humans are good at interpreting and answering questions based on such plots, VLMs often struggle~\citep{chartqa,yue2024mmmu}.

Just to give some examples of these requests, they may include tasks such as \textit{``report the highest tick on an axis''}, or \textit{``name the series with the largest gap''}, or \textit{``extract the value of a bar''} and more. It turns out that open-weight VLMs remain far below human accuracy on benchmarks built from real scientific figures~\citep{chartqa}.  Here, we use a variety of datasets: CharXiv, ChartQA, PlotQA, and EvoChartQA. CharXiv~\citep{charxiv} is a popular dataset curated by charts drawn from arXiv papers. It is reported that even strong models have a relatively low rate of correctness for answering descriptive questions compared to humans, and the gap widens on reasoning questions. ChartQA~\citep{chartqa}, PlotQA~\citep{plotqa}, and EvoChartQA~\citep{evochart} demonstrate a similar phenomenon across real-world, synthetic, and long-tail charts, with the gap being relatively largest for the small open models that are cheapest and easiest to deploy.

This work started from the failure modes of VLMs on such input data, and observing \emph{how} these models fail is quite instructive. For example, when asked for \textit{``the highest labeled $y$-tick on a chart''}, Qwen3-VL-2B~\citep{qwenvl} replied \textit{``The $y$-axis label is \dots\ the highest tick value is 58 \dots''}, even though the ground truth was \texttt{\textbf{60}}. Upon being asked for a percentage, the model responds, \textit{``The value is 28\%''}; however, the ground truth was \texttt{\textbf{38}}. As observed, the model fails to read the chart correctly in both cases, and since descriptive questions are graded by exact match, both answers are classified as wrong. 

More generally, inspecting the outputs of the baseline inference pipeline (Table~\ref{tab:qual}) shows that a large share of the ``errors'' are of similar flavor, i.e., the VLMs either provide incorrect values, or sometimes correct values may be present but wrapped in a heavily verbose sentence, followed by a unit or percent sign (or even preceded by the model's intermediate arithmetic). Perhaps surprisingly, even with a plain prompt and a generous token budget, Qwen3-VL-2B scores only 0.5\% on ChartQA descriptive questions and 0.0\% on PlotQA! These are observed to be failures of presentation and also perception.

The above observations raise the main driving question behind our work:

\begin{center}
\textbf{Main question:} \textit{How much accuracy in scientific chart understanding can we recover, without changing the model at all?}   
\end{center}

\paragraph{Baseline Inference Program.} Everything that lies between a benchmark question and the evaluated string is usually code wrapped around a fixed model (this is what we also refer to as the ``frozen'' model) consisting of the system prompt, per-question formatting hints, the decoding budget, and the post-processing that normalizes raw generation into an answer. From now on, we call this code the \emph{inference program}. In practice, it is often tuned by hand, one model and one benchmark at a time, and the results rarely transfer. We mention here that prompt optimizers such as OPRO~\citep{opro}, APE~\citep{ape}, and EvoPrompt~\citep{evoprompt} automate part of the work, but they search only over the prompt string. As a consequence, they do not increase accuracy for perception as much, since they cannot shrink a potentially verbose answer, nor can they strip a trailing percent sign, and so accuracy is being compromised. Of course, one could in principle do retraining on chart-specialized models~\citep{matcha,unichart}, but this is quite expensive, and in fact, impossible when weights of the model are closed.

\subsection{Our contribution}
\label{sec:contrib}

Our major finding is a proposed technique that completely avoids retraining, while significantly improving accuracy on all four benchmarks, thus answering the main question above affirmatively. The key technique is based on autonomous evolution, introduced originally for text-based Large Language Models (LLMs) by FunSearch~\citep{funsearch} and AlphaEvolve~\citep{alphaevolve}. 

Specifically, we can start with the initial inference program, which is a relatively small piece of Python code, and using autonomous evolution, we can improve it based on a cheap and reliable evaluator of the given responses. The edits are proposed by an LLM, which we call the \textit{LLM-judge} (in the experiments we used Gemini-2.5-flash). We apply this idea to VLMs and report an increased accuracy on the 4 tested benchmarks, for three diversely sized VLMs (up to 10B parameters).  

\begin{figure}[t]
\centering
\includegraphics[width=\linewidth]{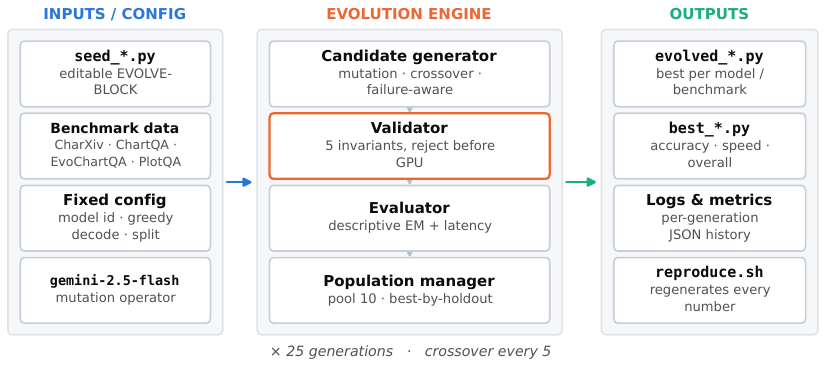}
\caption{\textbf{Our autonomous evolution framework for better VLMs.} Fixed inputs (left) feed an
evolution engine (middle) that outputs deployable inference programs
(right). The model id, greedy decoding, and the train/holdout split never
change during a run; only the code inside the \texttt{EVOLVE-BLOCK} of a
candidate is being mutated. An LLM mutator (in our case Gemini-2.5-flash) proposes edits, a validator rejects broken
candidate programs before any GPU work, an evaluator scores survivor programs by exact-match accuracy, and a population manager keeps the best-by-holdout program. The same engine produces all twelve runs by swapping the seed program and the benchmark evaluator.}
\label{fig:arch}
\end{figure}

\paragraph{Evolution of the inference program.} Fig.~\ref{fig:arch} shows how we evolve the inference program. Starting from the model as is (this is the ``frozen'' model), each new generation points the LLM-mutator to a few of the program's concrete failures and demands an improving edit in the inference code. The candidate programs live in a small pool ranked by training accuracy, and the next edit is made on the top-scorer. As is usual in such approaches, to limit overfitting, the final inference program that we output is the one that does best on a disjoint holdout window, that the search never optimized. To make this search practical, we use two safeguards: a static validator rejects malformed candidate programs before any GPU time is spent, and the model together with the VLM's decoding mode are fixed, so that every measured change in accuracy is attributable to the inference code (that we do the evolution on).

Below, we give more details on how we leverage the idea of autonomous evolution to improve (small) VLMs for the analysis of complex images like scientific charts.
\begin{itemize}
\item \textbf{Inference programs as the unit of optimization.} We formalize the executable inference program, instead of the prompt string or the model weights, as the object to be optimized for VLM chart question answering, and we build an evolutionary search framework over this space with an LLM as the mutation operator (Sec.~\ref{sec:method}).
\item \textbf{Safeguards that make the search trustworthy.} We introduce a validator that enforces five static invariants, and monitors GPU evaluation so that compute is spent only on valid candidates. This is then decoupled from a selection rule whose goal is to reduce overfitting. 

\item \textbf{Evaluation over 4 benchmarks and 3 models.} We finally evaluate on three differently sized open VLMs (Qwen3-VL-2B, InternVL3, SmolVLM2) and four chart benchmarks (CharXiv, ChartQA, EvoChartQA, PlotQA). We characterize when the method is helpful, when it is not, and what it costs (Sec.~\ref{sec:exp}).
\end{itemize}

\paragraph{Benefits of our proposed method.}
In comparison to simple prompt optimization, our search space contains the token budget of each response and the post-processing rule of the candidate inference program. By using the LLM mutator, and without the expensive retraining, we can perform a single edit to the inference program that later changes several of these together. In contrast to prompt optimization, our evolutionary search can fix errors that no re-wording could. Compared with retraining, the proposed method does not effectively change any weights, so it can be applied to any checkpoint that fits a ``one-function'' interface, including closed models. Comparing our method with hand-tuning, the search is automatic and reproducible: on Qwen3-VL-2B/ChartQA the evolved program matches a hand-tuned program in accuracy, and beats it on a combined accuracy-throughput score.

\begin{figure}[t]
\centering
\includegraphics[width=0.9\linewidth]{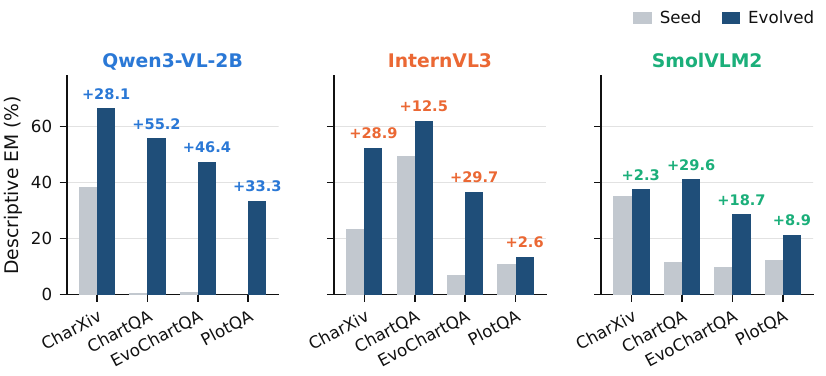}
\caption{\textbf{Seed vs.\ our evolved descriptive exact-match accuracy on all
12 model-benchmark pairs.} Annotations give the absolute gain in
percentage points by our method. As observed, every pair improves with the largest improvement being $+55.2$ percentage points. The largest jumps are on Qwen3-VL-2B, whose naive seed produced verbose, wrongly formatted
answers; the smallest is SmolVLM2/CharXiv, whose seed was already
clean (EM stands for Exact-Match).}
\label{fig:grid}
\end{figure}

\paragraph{Experimental results.}
Figure~\ref{fig:grid} and Table~\ref{tab:main} show a summary of the key results achieved by our proposed method. We report that across all twelve model-benchmark pairs, the evolved program improves descriptive exact-match accuracy over the native ``frozen'' baseline model. \textit{The average accuracy over the 12 combinations rises from 16.6\% to 41.3\%, and the relatively largest gain is 55.2 percentage points (Qwen3-VL-2B on ChartQA, from 0.5\% to 55.7\%).} There are several instances where more robust models' errors are perceptual (SmolVLM2 on CharXiv, $+2.3$; InternVL3 on PlotQA, $+2.6$). Upon analyzing the results of Qwen3-VL-2B/ChartQA, the fastest evolved variant is about 20 points more accurate than the unevolved model and 3.8 times faster, because the edits performed by our method produce short and concise answers, thus needing less compute. 

\begin{table}[t]
\centering
\small
\setlength{\tabcolsep}{5pt}
\begin{tabular}{@{}lrrrrrrrr@{}}
\toprule
& \multicolumn{2}{c}{CharXiv} & \multicolumn{2}{c}{ChartQA} & \multicolumn{2}{c}{EvoChartQA} & \multicolumn{2}{c}{PlotQA} \\
\cmidrule(lr){2-3}\cmidrule(lr){4-5}\cmidrule(lr){6-7}\cmidrule(lr){8-9}
Model & Seed & Evolved & Seed & Evolved & Seed & Evolved & Seed & Evolved \\
\midrule
Qwen3-VL-2B & 38.3 & \textbf{66.4} & 0.5 & \textbf{55.7} & 1.0 & \textbf{47.4} & 0.0 & \textbf{33.3} \\
InternVL3   & 23.4 & \textbf{52.3} & 49.5 & \textbf{62.0} & 6.8 & \textbf{36.5} & 10.9 & \textbf{13.5} \\
SmolVLM2    & 35.2 & \textbf{37.5} & 11.5 & \textbf{41.1} & 9.9 & \textbf{28.6} & 12.5 & \textbf{21.4} \\
\bottomrule
\end{tabular}
\caption{\textbf{Descriptive exact-match accuracy (\%) of the seed and the
evolved program for all 12 runs}, computed on the full descriptive
portion of each benchmark. Mean over runs: seed 16.6\%, evolved 41.3\%.}
\label{tab:main}
\end{table}

\section{Related Work}
\label{sec:related}

Prior work in the area of VLMs has addressed the performance gap between models and humans on chart question answering through different approaches, such as modifying the model weights, prompts, or inference programs. While these methods offer specialized optimizations, they either constrain the search space to text strings, operate within single-trajectory feedback loops or require modifying frozen model weights. In this section, we discuss these key research directions and how our work relates to them.

\paragraph{Chart understanding.}
Several benchmarks have been introduced for chart question answering, each focusing on different aspects of chart understanding.
Chart question answering was established as the benchmark task in ChartQA~\citep{chartqa}, while PlotQA~\citep{plotqa} introduced a large set of algorithmically generated questions, and EvoChartQA~\citep{evochart} expanded the space of question types in real-world charts. CharXiv~\citep{charxiv} further focused on targeted scientific charts from arXiv papers and separated \emph{descriptive} questions (evaluated via exact match) from open-ended \emph{reasoning} questions. The exact match protocol makes the formatting as important as semantic accuracy: correct values are marked incorrect if they are wrapped in sentences or extra units. Existing work improves chart understanding by changing the model, through chart-specific pretraining MatCha~\citep{matcha} and UniChart~\citep{unichart}, or through intermediate representations DePlot~\citep{deplot}, which converts a chart to a table before reasoning. These approaches require modifying the underlying model through retraining or fine-tuning. In contrast, our approach leaves the model weights untouched and modifies only the code around inference, including the prompt, decoding budget, and output parser. This makes the approach applicable to frozen models, including closed-weight models that cannot be modified through methods such as DePlot or MatCha.

\paragraph{Inference-time and prompt optimization.}
Prompt optimization methods such as OPRO~\citep{opro}, APE~\citep{ape}, and EvoPrompt~\citep{evoprompt} use an LLM to iteratively search over prompt strings, improving performance without model retraining. Since their search space is confined to prompt text, they cannot adjust decoding parameters, allocate dynamic token budgets, or add post-processing to normalize outputs into gradable answers. Other frameworks like DSPy~\citep{dspy} and TextGrad~\citep{textgrad} optimize declarative modules and text-valued parameters, and systems such as SGLang~\citep{sglang} and Ayo~\citep{ayo} optimize application execution around fixed weights. Building on this idea that the application layer is an ideal optimization target, we go one step further by searching directly over the executable program itself.

\paragraph{Evolutionary and self-improving code search.}
The idea of evolving executable code dates back to genetic programming~\citep{koza}, but it became more practical with the rise of LLMs. FunSearch~\citep{funsearch} showed that an LLM can serve as a mutator, discovering new constructions in combinatorics, and AlphaEvolve~\citep{alphaevolve} extended this idea to larger algorithmic and systems problems while maintaining population diversity. Our work builds on this key insight that operating over code, rather than prompt strings or model weights, allows the search to make structural changes that prompt-only optimization cannot express. In parallel, Self-Refine~\citep{selfrefine} and Reflexion~\citep{reflexion} showed that feedback can be used to improve outputs within a single trajectory, but they do not maintain a population or persist reusable code artifacts. Inspired by their use of concrete errors as feedback, we use a failure-aware mutation operator within a population-based framework. To our knowledge, this is the first work to apply LLM-driven code evolution to the inference pipelines of vision language models.

\paragraph{LLM serving systems.}
Existing LLM serving systems, such as Orca~\citep{orca}, vLLM~\citep{vllm}, and dLoRA~\citep{dlora}, boost throughput by low-level optimizations like request batching, memory management, and adapter-aware scheduling. Our work takes a complementary approach: instead of modifying the underlying serving infrastructure, we optimize what the model is asked to do, how much it may generate, and how its output is parsed. Since our optimizations focus on program structure rather than serving infrastructure, our evolved programs can be deployed seamlessly on top of any of these existing systems.

\section{Autonomous Evolution of VLMs}
\label{sec:method}
We start by giving a mathematical description for a fixed Vision Language Model and some notation used for the chart understanding task, the input question and the correct answer (we refer to it as the ``gold'' answer).
\subsection{Problem setting and key idea}
Let $f$ be a fixed VLM and let a benchmark be a set of triples $(x, q, a)$ of (chart image, question, gold answer). Each question carries an identifier for the type of question that it asks, labeled as $\mathit{qid}$ (for example, ``highest $y$-tick'' or ``number of series'' etc.). An \emph{inference program} $p$ is a function $p(x, q, \mathit{qid})$  mapping a specific triplet to a \text{string}. Function $p(\cdot)$ calls $f(\cdot)$ zero or more times, and returns a single string to be graded. Its descriptive accuracy in a set $S$ is measured by the exact match $\mathrm{EM}_S(p) = \frac{1}{|S|}\sum_{(x,q,a)\in S}\mathbb{P}[\,p(x,q,\mathit{qid}) = a\,]$ after normalization of the benchmark. We are given a training window $T$ and a disjoint holdout window $H$, and we seek a program that maximizes $\mathrm{EM}$ on questions it has never seen.

The key idea is that $p(\cdot)$ is a short Python module, and the undesired behavior that we want to fix is located within the program $p(\cdot)$ rather than in $f(\cdot)$. For example, a verbose answer that fails exact match is corrected not by re-wording the question, but by budgeting fewer tokens and by normalizing the raw string. Therefore, we make $p(\cdot)$ the object of our evolutionary search. Following techniques from FunSearch~\citep{funsearch} and AlphaEvolve~\citep{alphaevolve}, the search is an evolutionary loop in which an LLM is the so-called ``mutation'' operator, $\mathrm{EM}_T$ is the ``fitness,'' and a population of programs is maintained across generations. 

\subsection{Evolution search over candidate programs}
In the context of our evolution search, a candidate is a single Python module, where only the code between the markers \texttt{\# EVOLVE-BLOCK-START} and \texttt{\# EVOLVE-BLOCK-END} is being edited. Everything outside these markers is fixed, and this allows for a fast search. The changes are performed in one function, \texttt{vlm\_inference(image\_path, question, qid)} $\rightarrow$ \texttt{str}, which is responsible for mapping a plot and its question to an answer. Inside the editable code, our optimizer may change the system prompt, per-\emph{qid} formatting hints, per-\emph{qid} token budgets, and the post-processing that normalizes the raw output. Throughout our evolution search, both the model name and greedy decoding (\texttt{do\_sample=False}) are fixed by the validator and never change. We emphasize that the fixed decoding is what allows the generations to be reproducible, so that any accuracy difference between two generations is due to the code edit and not sampling noise (of the VLM).

\subsection{The evolution loop}
\begin{figure}[t]
\centering
\includegraphics[width=\linewidth]{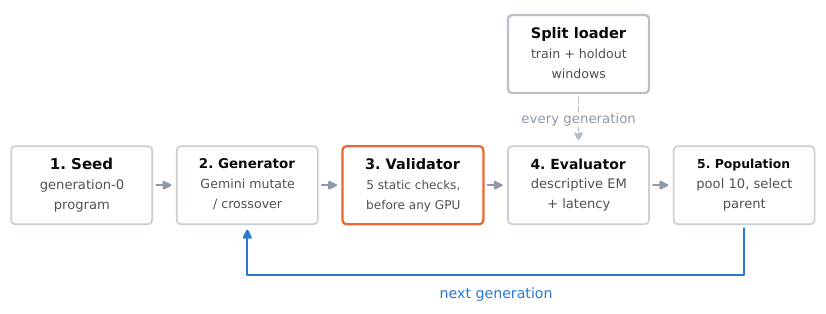}
\caption{\textbf{The evolution loop.} In every generation, the split loader feeds the train and holdout windows to the evaluator. The LLM-mutator proposes edits, the validator checks them before any GPU work is performed, and the population manager selects the next parent and writes the best-by-holdout candidate to disk. Every five generations, a crossover step combines two strong parents into one child.}
\label{fig:loop}
\end{figure}

In this section, we give descriptions of our evolution loop.
Each generation performs five steps as shown in Fig.~\ref{fig:loop}. The final outcome of this process is the best candidate inference program found during our search.

\paragraph{1. Seed.} Generation~0 is a deliberately naive baseline program: a plain prompt, a generous token budget, and no post-processing. Its accuracy is low by design, so that the search starts from the same place a practitioner would.

\paragraph{2. Candidate generator (mutation operator).}
The mutator is an LLM, Gemini~2.5~Flash~\citep{gemini}, in all experiments. As typical in evolution approaches, \emph{mutation} takes one strong parent and makes a single targeted edit, such as tightening the prompt, adjusting a per-\emph{qid} hint, changing a token budget, or rewriting post-processing. \emph{Crossover}, performed every five generations, takes two strong parents and synthesizes a child that keeps the better rule for each question type. A consequence of our proposed method is that the operator is \textit{failure-aware}: six concrete failures of the parent are sampled, and each triple (question, model answer, gold answer) is inserted into the prompt, so the LLM-mutator edits toward what is actually breaking rather than rewriting at random. The six are drawn from \emph{different} question types rather than being the six worst cases (we observed that concentrating on just one failing type would bias the mutator towards a narrow fix that only helps that question type at the expense of others).

\paragraph{3. Validation before compute.}
LLM-generated Python modules frequently break at run time through invalid chat formatting, incorrect imports that may fail, or false function signatures~\citep{tambon2025bugs,liu2023evalplus}. Because loading the VLM and running an evaluation is an expensive step, the validator's role is to reject bad candidate programs by performing cheap static checks before any GPU is used. Here, we enforce five invariants to check validity of Python modules (Table~\ref{tab:validator}). 

\begin{table}[t]
\centering
\small
\begin{tabular}{@{}rll@{}}
\toprule
\# & Invariant & Check \\
\midrule
1 & Markers & \texttt{EVOLVE-BLOCK} present \\
2 & Deterministic decode & no \texttt{do\_sample=True} \\
3 & Model unchanged & backbone id preserved \\
4 & Signature & \texttt{vlm\_inference(...)} intact \\
5 & Syntax & \texttt{ast.parse()} succeeds \\
\bottomrule
\end{tabular}
\caption{\textbf{The five validator invariants.} Checks 1--4 are cheap
string tests run before the costlier syntax and import stage; the whole
gate runs before any GPU work.}
\label{tab:validator}
\end{table}

\paragraph{4. Evaluator.}
The evaluator runs a top-scoring candidate program over the training window $T$ and the holdout window $H$ and reports four quantities. \emph{Descriptive exact match} is the primary signal and the only quantity that drives selection. \emph{Reasoning exact match} is computed as a proxy and is never used as fitness, because the official reasoning metric is open-ended and exact match is the wrong metric for it. In our experiments, we keep track of the \emph{latency} (hence throughput) and \emph{errors} (exceptions or empty generations), but these do not drive the search.

\paragraph{5. Population manager and selection.}
The population is a pool of at most ten programs sorted by their training accuracy. As usual in evolution search, the weakest of them are pruned in each generation, while parent selection is done via a tournament of three (80\% of the time) and a draw from the weaker half of the pool (20\% of the time), thus creating a balance between exploitation and exploration. Crucially, while parents are chosen based on \emph{training} accuracy, the candidate written to memory in each generation is the one with the best \emph{holdout} accuracy. Decoupling the two roles prevents overfitting. In our experiments, a run lasts 25 generations and produces the best-by-holdout program together with a JSON history of every event (see Appendix~\ref{app:logs}).

\section{Experiments}
\label{sec:exp}
In this section, we report the experimental setup (the code and files for which can be found at \url{https://github.com/AbhinavDubey30/VLM-Autonomous-Evolution}), observations and the experimental results obtained by the proposed technique to improve the VLMs. Our main result is that we significantly increased the VLM accuracy for each model and benchmark pair, while being faster than previous approaches (see Fig.~\ref{fig:pareto}).

\paragraph{Models.}
We evolve inference programs for three popular and compact open-source VLMs: Qwen3-VL-2B-Instruct~\citep{qwenvl}, InternVL3~\citep{internvl}, and SmolVLM2~\citep{smolvlm}. Qwen3-VL-2B is the strongest of the three on chart tasks; InternVL3 is included for its strong document and chart performance; SmolVLM2 represents a smaller, efficiency-oriented model. All three are run in float16 through HuggingFace Transformers~\citep{transformers} with greedy decoding.

\paragraph{Datasets.}
We use four extensive chart question-answering benchmarks: CharXiv~\citep{charxiv} (scientific charts from arXiv papers), ChartQA~\citep{chartqa} (real-world charts with human-written and machine-generated questions), EvoChartQA~\citep{evochart} (a wider range of question types), and PlotQA~\citep{plotqa} (large-scale algorithmically generated questions). Each benchmark's descriptive questions are split into a training window, used to compute fitness, and a disjoint holdout window, used only to choose the saved program. Reasoning questions are evaluated but never used for selection. For Qwen3-VL-2B/ChartQA the windows are 102 descriptive plus 51 reasoning training samples and 26 descriptive plus 13 reasoning holdout samples. Together this gives $3\times 4 = 12$ evolution runs.

\paragraph{Other methods.}
Our primary comparison is against the \emph{seed} program: baseline inference with no hints and no post-processing, evaluated end-to-end through the same pipeline as the evolved programs. For the Qwen3-VL-2B/ChartQA analysis, we add a hand-tuned program as a stronger human-designed baseline, and we report the three deployment variants that our pipeline extracts from the final pool (\texttt{best\_accuracy}, \texttt{best\_speed}, \texttt{best\_overall}). We do not run prompt-only optimizers such as OPRO or EvoPrompt as separate baselines; their search space is a strict subset of ours, and the qualitative analysis in Table~\ref{tab:qual} shows that the gains come predominantly from edits outside the prompt.

\paragraph{Experimental setup.}
The primary metric is descriptive exact-match accuracy after each benchmark's normalization. Speed is recorded in queries per second for every run but is not a selection signal. Each run performs 25 generations with a pool of at most ten programs, crossover every five generations, and up to six failure examples per mutation prompt; Gemini~2.5~Flash is the mutator throughout. Runs were performed on Google Colab A100 GPUs; one run takes on the order of a few hours on a single GPU. Final accuracies in Table~\ref{tab:main} are computed over the full descriptive portion of each benchmark, not only the holdout window.

\subsection{Results}
\label{sec:results}

\paragraph{Our evolution search improves every configuration.}
Table~\ref{tab:main} and Fig.~\ref{fig:grid} report baseline and evolved accuracy for all twelve runs. Mean accuracy rises from 16.6\% to 41.3\%, a mean gain of 24.7 points, and every run improves. The size of the gain depends strongly on the type of error the seed makes. Where the seed identifies the right answer but returns it in an ungradable form, the gain is large: Qwen3-VL-2B goes from 0.5\% to 55.7\% on ChartQA ($+55.2$) and from 1.0\% to 47.4\% on EvoChartQA ($+46.4$). Where the seed's outputs are already cleanly formatted, there is little room for the search: SmolVLM2 on CharXiv gains 2.3 points and InternVL3 on PlotQA gains 2.6. Ranking the twelve gains (Appendix~\ref{app:gains}) shows this split clearly.

\paragraph{Errors of the baseline models.}
Table~\ref{tab:qual} shows a few examples of incorrect responses of the baseline VLMs. In each, the seed's answer contains the incorrect value and is not gradable: it has a long explanation. In contrast, the evolved program returns the standardized string directly. These examples support the claim that the gain does not come only from better visual understanding; the search is improving the \textit{structure} of the inference program, and in particular how the answers are formatted and extracted.

\begin{table}[t]
\centering
\small
\setlength{\tabcolsep}{4pt}
\renewcommand{\arraystretch}{1.2}
\begin{tabular}{@{}p{0.31\linewidth}cp{0.50\linewidth}c@{}}
\toprule
\textbf{Question} & \textbf{Gold} & \textbf{Seed output} ($\times$) &
\textbf{Evolved} \tabularnewline
\midrule
\raggedright Highest labeled $y$-tick?
  & \textbf{60}
  & \raggedright ``The $y$-axis label is \dots\ the highest tick value is
    \underline{58} \dots''
  & \textbf{60} \tabularnewline
\addlinespace[3pt]
\raggedright Year of largest blue--green gap?
  & \textbf{2018}
  & \raggedright ``\dots\ we calculate: $2005\!:\,|43{-}46|{=}3$, $2016\dots$''
  & \textbf{2018} \tabularnewline
\addlinespace[3pt]
\raggedright \% supporting TikTok ban, Sept.
  & \textbf{38}
  & \raggedright ``\dots\ the value is \underline{28}. Therefore,
    \underline{28\%}''
  & \textbf{38} \tabularnewline
\addlinespace[3pt]
\raggedright \% with prior military service, 2023
  & \textbf{17}
  & \raggedright ``\underline{16\%}''
  & \textbf{17} \tabularnewline
\bottomrule
\end{tabular}
\caption{\textbf{Examples of incorrect instances (abbreviated).} The seed
baseline perceives incorrectly and often returns a long, verbose response; the evolved
program returns the exact answer. Wrong values in the seed output are
underlined. Examples are drawn from Qwen3-VL-2B, InternVL3 and SmolVLM2 runs.}
\label{tab:qual}
\end{table}

\begin{figure}[t]
\centering
\includegraphics[width=\linewidth]{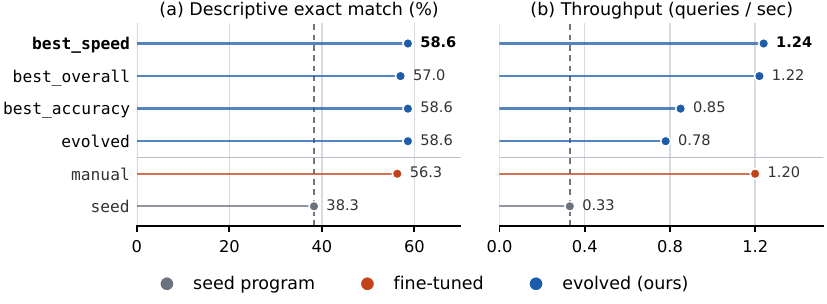}
\caption{\textbf{Accuracy vs.\ throughput, Qwen3-VL-2B/ChartQA}. Evolution moves the program up and to the right of the seed; \texttt{best\_speed} dominates on
both axes.}
\label{fig:pareto}
\end{figure}

\paragraph{Our evolution search is fast and accurate.}
Because the pipeline extracts variants tuned for accuracy, speed, and a balance of the two, we examine Qwen3-VL-2B on ChartQA in detail. Table~\ref{tab:deepdive} and Fig.~\ref{fig:pareto} report all six programs on a window of 128 descriptive and 64 reasoning samples, together with an aggregate score, the harmonic mean of descriptive accuracy and throughput capped at 10 queries per second, floored to zero below 30\% accuracy (Appendix~\ref{app:score}). This score is used only for ranking deployment variants; the search itself optimizes descriptive exact match alone. On this window, the seed reaches 38.3\% at 0.33 queries per second. In contrast, our \texttt{best\_speed} variant reaches a better accuracy of 58.6\%, while operating at a higher throughput of 1.24 queries per second. This is about 20 points more accurate and 3.8$\times$ faster, thus making for an improved VLM overall. 

Moreover, even though one might expect exact-match gains to come at a cost for latency, through longer prompts or more deliberate reasoning, we observe the opposite behavior. This is because the edits performed by our evolution approach create brief answers that are more easily gradable, and also remove unnecessary generation. Thus, they are overall much faster to produce.  

\begin{table}[t]
\centering
\small
\setlength{\tabcolsep}{3.5pt}
\begin{tabular}{@{}lrrrr@{}}
\toprule
Program & Desc.\ EM & Rsn.\ EM & QPS & Overall \\
\midrule
seed (starting)   & 38.3\% & 4.7\%  & 0.33 & 0.061 \\
hand-tuned (human) & 56.3\% & 17.2\% & 1.20 & 0.198 \\
evolved           & 58.6\% & 9.4\%  & 0.78 & 0.138 \\
best\_accuracy    & 58.6\% & 10.9\% & 0.85 & 0.148 \\
\textbf{best\_speed} & \textbf{58.6\%} & 14.1\% & \textbf{1.24} & \textbf{0.205} \\
best\_overall     & 57.0\% & 9.4\%  & 1.22 & 0.201 \\
\bottomrule
\end{tabular}
\caption{\textbf{Qwen3-VL-2B / ChartQA, all six programs} (128 descriptive and
64 reasoning samples). \texttt{best\_speed} is both the fastest and the
highest on the aggregate score, beating the hand-tuned baseline.}
\label{tab:deepdive}
\end{table}

\section{Conclusion}
\label{sec:conclusion}

In this study, we demonstrated that VLM inference for scientific chart question answering can be substantially improved by proposing an evolution-based method that edits the inference program rather than the prompt alone. For three different VLMs, we found improvements across four benchmarks without retraining the models. For all 12 model-benchmark pairs, we improved the mean descriptive exact-match accuracy from 16.6\% to 41.3\%, with the largest gain of 55.2 points. The method combines an LLM-based mutation operator, GPU evaluation gated by a static validator, and a best-by-holdout selection rule. The fastest evolved program on Qwen3-VL-2B/ChartQA was 3.8 times faster than the seed baseline program and simultaneously more accurate than a hand-tuned baseline. We observe that our method performs most effectively when the correct answer is wrapped in verbose reasoning in the model output. We observed that it was relatively less effective with perceptual failures and in cases with no model response. 

\section*{AI Use Statement}
In this work, we used generative AI tools to implement methods, in the specific sense that Gemini~2.5~Flash is the mutation operator of our method and writes the candidate inference programs that we evaluate (Sec.~\ref{sec:method}); this use is the subject of the paper, every program the model produced was validated automatically and evaluated empirically, and the evolved programs are research artifacts of this work. We have not used generative AI tools to develop theoretical models or conceptual frameworks, to propose or refine hypotheses, to design or provide feedback on research methodology or experiments, to interpret results, or to support qualitative or thematic data analysis. Generating synthetic datasets, formulating or proving mathematical claims, translation, and cleaning or reformatting datasets are not applicable to this work. Additionally, we used generative AI tools to edit the paper to improve readability, to suggest a structure for the paper, to format references, and to check compliance with the formatting instructions. We have reviewed all AI-assisted work. We take responsibility for the final content of this work, including text, claims or artifacts produced with the aid of generative AI.

\section*{Reproducibility Statement}
All runs use greedy decoding with the model id fixed by the validator, so re-running a saved program reproduces its outputs exactly. Sec.~\ref{sec:method} specifies the candidate contract, the validator invariants (Table~\ref{tab:validator}), the mutation and crossover operators, and the selection rule; Sec.~\ref{sec:exp} specifies the models, benchmarks, split sizes, generation count, pool size, crossover interval, and hardware. Appendix~\ref{app:logs} describes the code layout and the per-generation JSON history, and Appendix~\ref{app:score} defines the aggregate score used in the deep dive. The source code, seed programs, evolved programs, and evolution logs for all twelve runs can be found at \url{https://github.com/AbhinavDubey30/VLM-Autonomous-Evolution}.

\bibliography{iclr2027_conference}
\bibliographystyle{iclr2027_conference}

\appendix

\section{Aggregate score for the deep dive}
\label{app:score}
For ranking the deployment variants in Table~\ref{tab:deepdive} we use
\[
\textsc{overall}=\mathrm{harmonic\_mean}(\text{desc\_acc},\ \min(\text{QPS}/10,\,1)),
\]
set to zero when descriptive accuracy falls below 30\%. The harmonic mean penalizes programs that are strong on only one dimension: a program should not rank highly if it is fast but mostly wrong, or accurate but too slow to be useful. The 30\% floor reflects the same practical concern, since a chart reader with very low accuracy has little deployment value regardless of speed. The score is used only for reporting; the evolutionary search optimizes descriptive exact match alone.

\section{Code structure and logging}
\label{app:logs}
The implementation is roughly six Python files, organized so that the engine is reused unchanged across runs:
\begin{itemize}
\item \texttt{seed\_*.py}: the generation-0 program, naive inference with no hints or post-processing.
\item \texttt{evolve\_<benchmark>.py}: the engine, comprising the population manager, parent selection, mutation and crossover, validation, and the generation loop.
\item \texttt{evolve\_common.py}: validation helpers, including reading the required model name from the seed and rejecting any model swap before GPU time.
\item \texttt{evaluate.py} / \texttt{evaluate\_<benchmark>.py}: load images and answers, run a candidate, compute accuracy and timing.
\item \texttt{evolved\_<benchmark>.py}: the winner after 25 generations, selected on questions it never saw during evolution.
\item \texttt{best\_accuracy.py}, \texttt{best\_speed.py}, \texttt{best\_overall.py}: three deployment variants drawn from the final pool.
\end{itemize}
Inference uses PyTorch with HuggingFace Transformers~\citep{transformers} (\texttt{AutoModelForImageTextToText} / \texttt{AutoProcessor}) in float16 single-query mode. The mutator is called through the Google Generative AI SDK. Every generation prints the parent, the child's train and holdout accuracy, and the current best, and a JSON history records each event (seed, improvement, no\_improvement, crossover, mutation\_failed) with accuracies and timing. A representative excerpt:

\begin{lstlisting}[language=,basicstyle=\ttfamily\scriptsize]
[Gen 7/25] Parent: 56.9%/57.7% (gen 5)
  Child: train=58.6% holdout=61.5%
  Best by holdout: 61.5% -> saved
[Gen 10/25] [CROSSOVER] A:58.6/61.5
                        B:55.5/57.7
  Child: train=57.0% holdout=61.5%
  (best-by-train != best-by-holdout)
\end{lstlisting}

\section{Per-run gains, sorted}
\label{app:gains}
\begin{figure}[t]
\centering
\includegraphics[width=0.8\linewidth]{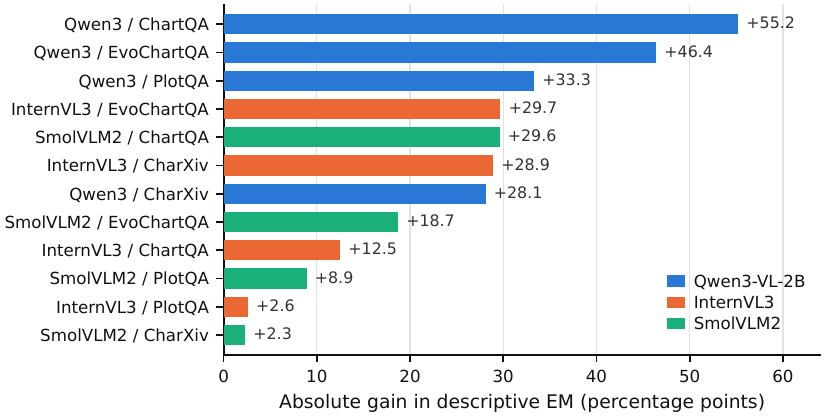}
\caption{\textbf{Absolute gain in descriptive exact match for each of the 12
runs, sorted.} Color denotes model. Gains cluster by whether the seed's
failure was one of \emph{formatting} (large, recoverable) or
\emph{perception} (small headroom).}
\label{fig:gains}
\end{figure}

Figure~\ref{fig:gains} ranks the twelve runs by their absolute gain in descriptive exact match.

\section{Engineering frictions}
\label{app:frictions}
Several candidate models were excluded for practical reasons. Some gated checkpoints, such as MiniCPM-V-2.6 and PaliGemma2, were delayed by access approvals. Others, including Molmo, Pixtral, and Phi-3.5-Vision, did not load correctly under our unified inference template. We also tested and discarded two efficiency strategies. Request batching for Qwen3-VL-2B ran into a right-padding issue in its \texttt{Qwen2VLProcessor} that capped accuracy at 29.7\%. INT4 quantization ran at only 0.04 queries per second on a T4 GPU, roughly 75 times slower than float16, and was not pursued. These are not algorithmic failures, but they reflect the practical difficulty of building a uniform evaluation harness across heterogeneous VLMs.

\end{document}